# Challenges in Persian Electronic Text Analysis

Behrang QasemiZadeh [a][1], Saeed Rahimi [b] and Mehdi Safaee Ghalati [b]

[a] *Text and Speech Technology Ltd., Tehran, Iran*
[b] *Tehran University, Faculty of Literature and Humanities, Enqelab, Tehran, Iran*

Farsi, also known as Persian, is the official language of Iran and Tajikistan and one of the two main languages spoken in Afghanistan. Farsi enjoys a unified Arabic script as its writing system. In this paper we briefly introduce the writing standards of Farsi and highlight problems one would face when analyzing Farsi electronic texts, especially during development of Farsi corpora regarding to transcription and encoding of Farsi e-texts. The pointes mentioned may sounds easy but they are crucial when developing and processing written corpora of Farsi.

## 1 INTRODUCTION

People in different countries use different characters to represent the words of their native languages. With library automation and the development of networked information structures, the problem of finding a unique way to show information has become much more complex [1][2]. Unicode [4] was devised so that one unique code is used to represent each character, even if that character is used in multiple languages [3]. In this paper, we describe Farsi language transcription in Unicode framework and we discuss challenges that someone would face when processing Farsi e-texts.

Old Farsi was based on the cuneiform writing system (a mostly syllabic style) as early as the 6th century B.C. Later, the Persians developed a new alphabet called Pahlavi, which was derived from Aramaic writing system, a Semitic language, to replace the previous uniform alphabet. However, after the Arab's conquest in 651, the Persians adopted a unified Arabic script for writing. One should note that despite their shared alphabet, Farsi and Arabic are entirely different languages: they are not genealogically related (they belong to separate language families, named, Indo-European and Afro-Asiatic) and naturally have different phonology and grammar [5] [6].

With the expansion of Islam, Arabic script was adopted as a system of writing also for other languages other than Arabic. As in many of these languages, among them Farsi, Urdu, and Sindhi, a greater number of phonemes, compare to the Arabic language, had to be depicted in written form, the repertoire of Arabic characters was extended. The original Arabic language alphabet consists of 28 characters. The modern Farsi writing system uses the Arabic alphabet, but with the addition of four letters which do not occur in Arabic. These are:" گ ژ پ چ ". Additionally, it changes the shape of another two i.e. "ی" and "ک". Not all of the sounds represented in the Arabic alphabet exist in Farsi; as a result, more than one letter may represent one sound. For example, there are two letters in Farsi for the sound /t/ (ط ت) and three for the sound /s/ (ث ص س).

Salient characteristics of Arabic script are: existence of various connecting letters, varying graphic forms for many letters depending on their position in a word, varying letter width, absence of full size characters for vowels (vowels are represented with particular signs above and below characters), existence of a number of digraphs and composite letters, writing direction from right to left and absence of upper case and lower case letters. General rules of Arabic writing system are followed by the writing system of Farsi.

In the following, we will introduce writing system of Farsi in detail and we will underline problems of this writing system when analyzing Farsi e-texts. The rest of paper is organized as follows: section 2 introduces Farsi character encoding. Section 3, describes the orthography of Farsi. Section 4 gives an overview of common ambiguities in the analysis of Farsi e-texts. Finally, we conclude in section 6.

---

[1] Corresponding Author: Behrang Qasemi Zadeh, qasemizadeh@gmail.com

## 2  FARSI CHARACTER ENCODING

The definition and universal implementation of character sets for the presentation, interpretation, and exchange of multi script data is a problem ever since computers were adapted for Information Retrieval applications. As for the Farsi script, various attempts have been done in the past to create a universally acceptable and technically viable encoding system. The most prominent result was the *Iran System*. Note that *Iran System* is not a standard; it is a corporate character set that found its way through the Iranian user community. Therefore, there is no standardization paper to refer to. The characters in the *Iran System* standard are saved, and transmitted in a visual order. There are two, three, or four codes for each letter in most cases: for initial, medial, isolated, and final forms of the letters, some of which are unified in most of the cases. In other words, in character encoding system before Unicode, glyphs are encoded instead of conceptual letters. After a while a standard for 8-bit Farsi character encoding was proposed in [7] by the Institute of Standards & Industrial Research of Iran but it did not catch on with user community.

After a while, The ISIRI 6219:2002 (Information Technology – Persian Information Interchange and Display Mechanism, using Unicode) was proposed as Farsi standard for using Unicode in digital environment. This standard indicates a subset of Arabic character set in Unicode to be used by user communities for Farsi. In this document, we will refer to ISIRI 6219:2002 as Farsi Standard Character set.

Unicode standard version 4.0 reserves the range 0600 to 06FF for Arabic characters. Among the 227 Arabic script signs currently encoded, there are punctuation marks, pronunciation marks, symbols for honorifics and Koran's annotation signs as well as all letters representing consonants in Arabic and the other languages using Arabic script. Important design principles observed in the Unicode standard and relevant to the Representation of Arabic script are characters not glyphs. Some Arabic letters can have up to four different positional forms depending on their position relative to other letters or spaces. According to the design principle "characters, not glyphs" there is no individual code for each visual form (glyph) that an Arabic character can take in varying contexts but only one code for each actual letter. The correct glyphs to be displayed for a particular sequence of Arabic characters can be determined by an algorithm. Encoding algorithm for Arabic transcribed texts uses "first-to-last" logical order instead of "right-to-left" or "left-to-right" for correct representation in cases a text contains "right-to-left" and "left-to-right" strings.

As mentioned earlier, an algorithm is responsible for displaying correct glyphs of characters. For this reason, characters are classified according to their ability to join their previous or next characters onto different shaping classes. For proper displaying of characters two special characters namely Zero Width Joiner (0x200D) and Zero Width Non Joiner (0x200C) are added to character codes. The use of these special characters after a code means that a ZWJ or a ZWNJ should be added after the character if the character is not followed by a "right-join causing" character, or a "non-joining character" respectively. Unfortunately, even with the developed standard for using these characters in Farsi electronic texts, the user community does not respect it; thus, this is caused ambiguity in Farsi e-texts manipulation as we talk about it later.

## 3  FARSI ORTHOGRAPHY

Iran's Academy of Farsi Language and Literature[2] is a governmental body presiding over the use of Farsi in Iran. The Academy has created an official orthography of Farsi, entitled 'Dastoor-e Khatt-e Faarsi' (Farsi Script Orthography). Official orthography of Farsi can be found in [9]. According to the proposed orthography, Farsi affixes must be written in an attached form to their stem. In some cases e.g. when the stem end in letter "ه"(h), affixes must be attached to the stem with a short space character before them. Here, we take ZWNJ character as the short space character. Most of the ambiguities in Farsi morphology and Part of Speech tagging of Farsi e-texts will be avoided using the standard that has been proposed by the academy as discussed in the next section.

---

[2] http://www.persianacademy.ir

## 4 FREQUENT AMBIGUITIES IN ANALYSIS OF FARSI

When starting the computational analysis of Farsi, one would face a lot of ambiguities which root in Farsi language characteristics and its special transcription. In this section we have discussed these ambiguities and we propose our solutions for some of these problems.

### 4.1 Ambiguities while Characters Manipulation

In Farsi, the use of Arabic characters instead of Farsi standard ones is possible. A common mistake is associated with letters "ى" and "ي" as well as "ك" and "ک". Problems arise when using dictionaries for looking up words, making frequency profiles of words due to the inconsistency in encoding. Even this problem causes different results when using keyword based search engines like Google.

Similar problem happens when a short vowel enters in a word transcription. Because short vowels do not appear alone in Farsi transcriptions but when using these short vowels in a word, they are coded independently. Therefore, there is a difference in Unicode strings of the same word with or without short vowels included. This can lead to an unsuccessful search in a dictionary or lexicon.

Another problem concerns the character "TATWEEL". As we mentioned in the characteristics of Arabic transcription, letters can appear with varying width. This is just a visual characteristic and it does not influence the meaning of a word. "TATWEEL" character with code (0640) is used to support this characteristic. We have to consider that this character must be deleted from input string when looking up dictionaries etc.

To solve this problem a standardization procedure must be undertaken for input Farsi e-texts. This standardization procedure can vary from one application to another. But using a standard character set for constant letters is obvious. This can be done using a map between Arabic characters to Farsi ones. For some cases, such as keyword based searching, it is recommended to omit short vowels from input to have a more consistent search, it is clear that the omitting of short vowels results in loss of information. Even these vowels can be used as signs for solving the problem of homographs in Farsi transcription. We recommend omitting these vowels from input texts even at the cost of losing information due to the fact that using short vowels in Farsi is rare but actually it is much dependant on the context.

### 4.2 Ambiguity at Word Boundaries

In Farsi, word boundaries can be delimited by space, punctuation, and the forms of the characters indicating their position within a word. Word boundaries are usually denoted by space. Considering the official orthography of Farsi, space is an unambiguous word boundary. For some cases such as compound words and light verb constructions ZWNJ is used for separating their different parts. Unfortunately, the user communities do not consider this; even the organizations like newspapers. For this reason in a normal text, space cannot be considered as an unambiguous word boundary and vice versa. Here there appears a conflict between ZWNJ and Space character that should be considered.

Full stop marks a sentence boundary, but it may also appear in the formation of abbreviations or acronyms. The slash (/) is used in the numbers and dates structures. Also the dash (-) could be used to separate compound words. Other punctuation marks including the comma, quotes, brackets, question mark and colon unambiguously indicate word boundaries.

As mentioned above, Arabic characters take one of the four forms: initial, medial, final, and stand-alone, and the final form of characters may indicate the end of a word. Character form can be used as a delimiter depending on the encoding structure. This condition is not applicable when using Unicode as an encoding system. This is a common mistake associated with tokenization in Farsi, as you can see in [10][11]. This condition can be used for word boundary detection when using old encoding systems because they use different character codes for different glyphs of a letter. Even while using an old encoding system of Farsi this condition can raise ambiguity as you can see in [11]. One may argue that by using ZWNJ as an end of word marker, we can overcome this problem but it cannot be right considering the official orthography of the Farsi.

### 4.3 Ambiguity in morphology

Ambiguity in morphological analysis of words in Farsi arises for two reasons. One for homograph words and the other is the ambiguity caused by word boundary. According to [11] in Farsi, a single surface form can represent different morphemes. In addition, short vowels are not marked in written texts, which results in different possibilities for analysis. The word ببر(*bbr*), for instance, can be pronounced with different vowel combinations resulting in three possible common lexical elements: ببر(babr) which means "tiger", ببر(bebar) which means "take" and ببر(bebor) which means "cut".

As we mentioned in 5.3, user communities use space character instead of ZWNJ, short space. For this reason, certain bound affixes appear as free morphemes. For example, the affix "ها" (ha) can be written in three different ways as shown in the example below with the same meaning:
– As bound morpheme "کتابها" (keta<u>b</u><u>h</u>a) which means "books" in English.
– As free morpheme with a space between the root and bound morpheme "کتاب ها" (ketab ha).
– As free morpheme with a short space(ZWNJ) instead of a space character کتاب‌ها"

The last format is the right orthography according to [9].

Similar problem usually happens for some other lexical elements, such as the preposition "به" (be), the postposition (object overt marker) "را" (ra), or the conjunction "که" (ke) that usually appear as separate words in written texts, but can also be found as attached morphemes. By using official orthography this lexical elements must be written separately. Another solution for detached bound morphemes is the morpheme-based processing of Farsi written texts.

### 4.4 Ambiguity while Detecting Proper Nouns in Farsi

Since there is no capital letter in Arabic transcription and as a result in Farsi transcription, detecting proper nouns in Farsi brings some problems. In Farsi, there is no general rule for distinguishing proper names from the other nouns. However, some heuristics can be used to distinguish proper names. Rezaie addressed a solution for this problem in [12].

### 4.5 Ambiguity in Farsi Syntax analysis

Another ambiguity arises in possessive constructions in Farsi due to its Arabic transcription. The element joining the Farsi noun phrase constituents to each other is the *Ezafe* clitic. However, it is usually pronounced as the short vowel /e/ and therefore is not marked in written texts. The result, in Farsi written texts, is a series of consecutive nouns without any overt links or boundaries. In [9] it is recommended to include *Ezafe* in written texts but as we mentioned before in section 4.1 this can result in some ambiguities while manipulating Farsi texts.

## 5 CONCLUSION

In this paper, we discussed problems that may occur when analyzing Farsi e-texts due to the inconsistency of representing Farsi character and its special orthography. Although the points we have mentioned here, may sound trivial but not considering them may lead to wrong results, which are far away from real situation.

Our study shows that the combination of Farsi orthography that is proposed in [9] and the standard in [8] is the best way to remove ambiguities associated with Farsi e-texts. Considering these standards will result in consistency between Farsi and other languages especially developing parallel corpora. Still we have to find a solution for putting the existing Farsi texts in these standards. As a step to this goal, we started to develop a standardization tool to cope with the mentioned problems. We hope that we can put it as an online tool for user community of Farsi. In addition, it is needed that software community respect these standards especially the keyboard layouts they prepare must be compatible with the proposed standards in [8]. Also according to the importance of ZWNJ character in Farsi transcription it is needed to put this character in keyboard layout in a way that it can be easily available for users.